\documentclass[sigconf,authorversion,nonacm]{acmart}
\usepackage{graphicx}
\usepackage{xcolor}
\definecolor{darkgreen}{HTML}{7aeb34} 
\definecolor{darkred}{HTML}{eb3453}   
\usepackage[xindy, acronym, nohypertypes={acronym}]{glossaries-extra}
\usepackage{xspace}
\usepackage{amsmath}
\usepackage{wrapfig}
\usepackage{colortbl} 
\usepackage{enumitem}
\usepackage{makecell}
\usepackage{hyperref}

\newcommand{\refSec}[1]{Sec.~\ref{sec:#1}}
\newcommand{\refFig}[1]{Fig.~\ref{fig:#1}}

\newcommand{\refEq}[1]{Eq.~(\ref{eq:#1})}
\newcommand{\refTbl}[1]{Tbl.~\ref{tbl:#1}}

\AtBeginDocument{%
  }

\begin{document}
\newabbreviation{FPS}{FPS}{Frames Per Second}
\newcommand{\FPS}{\gls{FPS}\xspace}

\newabbreviation{CGH}{CGH}{Computer-Generated Holography}
\newcommand{\CGH}{\gls{CGH}\xspace}

\newabbreviation{SLM}{SLM}{Spatial Light Modulator}
\newcommand{\SLM}{\gls{SLM}\xspace}

\newabbreviation{VAC}{VAC}{Vergence-Accommodation Conflict}
\newcommand{\VAC}{\gls{VAC}\xspace}

\newabbreviation{DOE}{DOE}{Defractive Optical Element}
\newcommand{\DOE}{\gls{DOE}\xspace}

\newabbreviation{INR}{INR}{Implicit Neural Representation}
\newcommand{\INR}{\gls{INR}\xspace}

\newabbreviation{NeRF}{NeRF}{Neural Radiance Field}
\newcommand{\NeRF}{\gls{NeRF}\xspace}

\newabbreviation{PSF}{PSF}{Point Spread Function}
\newcommand{\PSF}{\gls{PSF}\xspace}

\newabbreviation{DFT}{DFT}{Discrete Fourier Transform}
\newcommand{\DFT}{\gls{DFT}\xspace}

\newabbreviation{IDFT}{IDFT}{Inverse DFT}
\newcommand{\IDFT}{\gls{IDFT}\xspace}

\newabbreviation{MLP}{MLP}{Multi-Layer Perceptron}
\newcommand{\MLP}{\gls{MLP}\xspace}

\newabbreviation{MSE}{MSE}{Mean Square Error}
\newcommand{\MSE}{\gls{MSE}\xspace}

\newabbreviation{DDM}{DDM}{Denosing Diffusion Model}
\newcommand{\DDM}{\gls{DDM}\xspace}

\newabbreviation{NSR}{NSR}{Noise-to-Signal Ratio}
\newcommand{\NSR}{\gls{NSR}\xspace}

\newabbreviation{LCD}{LCD}{Liquid Crystal Display}
\newcommand{\LCD}{\gls{LCD}\xspace}

\newabbreviation{ROI}{ROI}{Region Of Interest}
\newcommand{\ROI}{\gls{ROI}\xspace}

\newabbreviation{FPD}{FPD}{Flat Panel Displays}
\newcommand{\FPD}{\gls{FPD}\xspace}

\newabbreviation{OST-HMD}{OST-HMD}{Optical See-Through Head-Mounted Displays}
\newcommand{\OSTHMD}{\gls{OST-HMD}\xspace}

\newabbreviation{3D}{3D}{Three-Dimensional}
\newcommand{\threeD}{\gls{3D}\xspace}
\title{Learned Display Radiance Fields with Lensless Cameras} 
\author{Ziyang Chen}
\affiliation{%
  \institution{University College London}
  \city{London}
  \country{United Kingdom}
}
\email{ziyang.chen.22@ucl.ac.uk}

\author{Yuta Itoh}
\affiliation{
  \institution{Institute of Science Tokyo}
  \city{Tokyo}
  \country{Japan}
}
\email{itoh@comp.isct.ac.jp}

\author{Kaan Akşit}
\affiliation{
  \institution{University College London}
  \city{London}
  \country{United Kingdom}
}
\email{k.aksit@ucl.ac.uk}

\begin{abstract}
Calibrating displays is a basic and regular task that content creators must perform to maintain optimal visual experience, yet it remains a troublesome issue.
Measuring display characteristics from different viewpoints often requires bulky equipment and a dark room, making it inaccessible to most users.
To avoid such hardware requirements in display calibrations, our work co-designs a lensless camera and an Implicit Neural Representation based algorithm for capturing display characteristics from various viewpoints.
More specifically, our pipeline enables efficient reconstruction of light fields emitted from a display from a viewing cone of $46.6^\circ \times 37.6 ^\circ$.
Our emerging pipeline paves the initial steps towards effortless display calibration and characterization.
\end{abstract}


\begin{teaserfigure}
  \includegraphics[width=\textwidth]{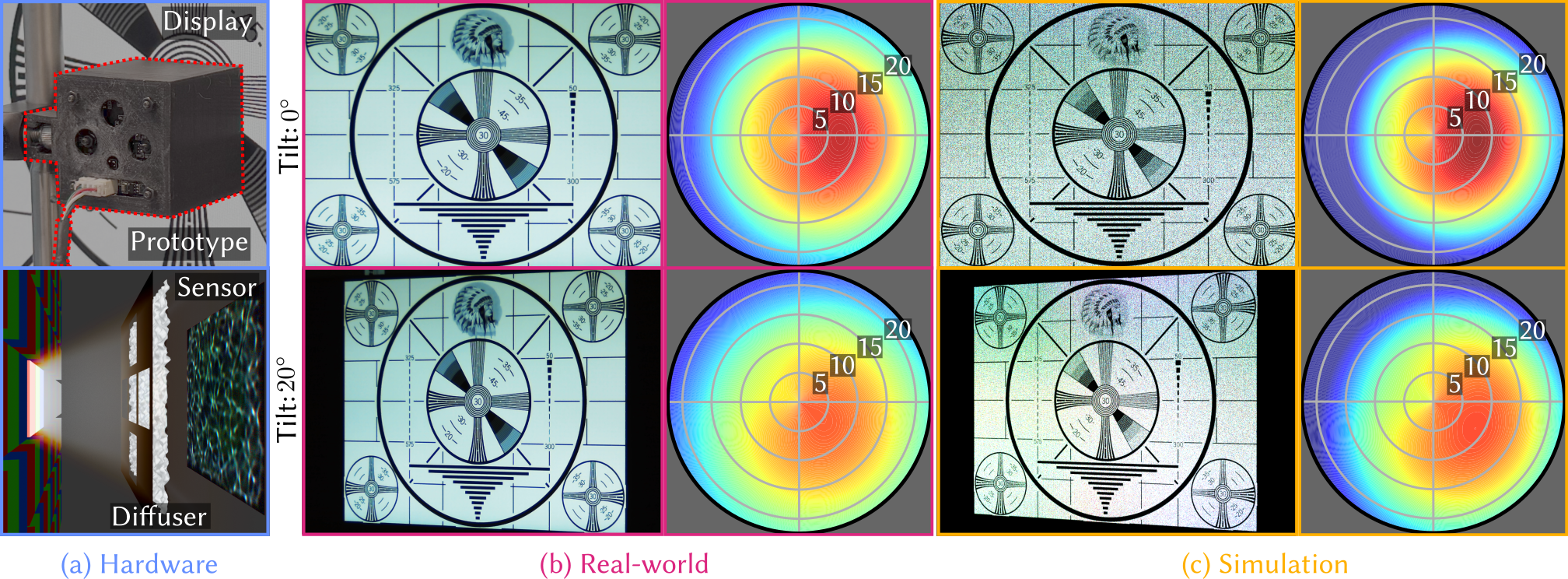}
  \caption{
  (a) Our lensless camera placed in front of the display captures the light field emitted from the display pixels. 
  (b) A conventional camera captures a real-world photograph of the test pattern on the display. 
  (c) Our learned model estimates a rendered view from the viewpoint of the real-world photograph displayed in the middle column. 
  The right columns in (b) and (c) present the display's angular intensity distributions in spherical coordinates. 
  The radius denotes the combined angular deviation from the optical axis, computed from the horizontal and vertical incidence angles.
  These plots illustrate how the relative intensity changes with viewing angle.
  }
  \label{fig:teaser}
\end{teaserfigure}

\maketitle
\section{Introduction}

Professional content creators use off-the-shelf colorimeters~\footnote{https://www.adobe.com/creativecloud/video/discover/how-to-calibrate-monitor.html} regularly to maintain a color consistency in the workflow.
Similarly, display engineers conduct calibrations by measuring various aspects of screens including but not limited to luminance, chromaticity, and contrast ratios~\footnote{https://www.admesy.kr/articles/display-calibration-workflows-rd-to-mass-production/}.
This process requires specialized equipment such as spectroradiometers or goniophotometers~\footnote{https://www.lisungroup.com/news/technology-news/application-of-goniophotometers-in-display-and-projection-technologies.html}.


Display chromaticity and luminance vary with viewing angle, affecting how users perceive content from different positions in front of a screen.
But most calibration tools assume a fixed viewing position leading to invalid assessments for other viewing positions.
This is not only problematic for display manufacturers but also for professionals that need consistent color edits for various geometric configurations.
The ISO standards~\cite{ISO9241-305} define procedures for assessing the uniformity distribution of \FPD at various viewing directions by rotating either the display or measurement instrument in a dark room.
However, these requirements with numerous captures hinder the end users from performing regular calibration, which is critical for virtual production display walls~\footnote{https://partnerhelp.netflixstudios.com/hc/en-us/articles/1500002086641-Common-Virtual-Production-Challenges-Potential-Solutions}, professional graphic design, and video editing~\footnote{https://displaycal.net/}.
\textit{An accessible method with a reasonable amount of captures is needed to measure the image quality of the displays from arbitrary viewpoints.}

Our work introduces a calibration tool that characterizes the display pixel emission patterns at different viewing angles with simpler setup and fewer captures. 
Specifically, we design a paired hardware and software: (1) a lensless camera~\cite{chen2024spectrack, kingshott2022unrolled} capturing the incoming light of a display pixel from a range of directions and (2) an \INR that encodes the display angular responses. 
Our work makes the following contributions:
\begin{itemize}[leftmargin=*]
  \item \textbf{Lensless Camera Prototype}.
  We design a lensless camera using a phase mask and aperture array to capture display characteristics across a practical angular range.  
  The phase mask is positioned $30\,mm$ from the display pixel, with the imaging sensor placed $10\,mm$ behind it to ensure spatial invariance of the diffuser.  
  This configuration enables horizontal and vertical incident angle coverage of approximately \textbf{46.6} and \textbf{37.6} degrees, respectively.  
  \item \textbf{\INR for Display Pixel Characterization}.
  To efficiently represent the light fields from the lensless captures and generate novel views of unseen display pixels, we train an \MLP that performs end-to-end reconstruction based on different display positions and incident angles.
  This \MLP model can be trained with only \textbf{nine} positions on the display, enabling the reconstruction of the light field for each pixel on the display.
\end{itemize}
We conduct experiments on a \LCD and demonstrate that our method can reconstruct the display with different viewpoints.
Our codebase is available at \href{https://github.com/complight/learned_display_radiance_fields_with_lensless_cameras}{GitHub:complight}.

\section{Methods}
Our method reconstructs the light field representation of a display from lensless captures of individual pixels.
We detail our pipeline consisting of hardware and software components in the subsequent sections.
Our notations are summarized in \refTbl{notation}. 
Upper-case bold letters denote matrices, while lower-case bold letters represent vectors. 
Scalars and spatial indices are denoted by non-boldface letters. 
The calligraphic letters indicate the functions. 

\begin{table}[h]
\centering
\begin{tabular}{ll}
\hline
\( \mathbf{H} \) & Coefficient matrix for lensless forward model \\
\( \mathbf{Y} \) & Lensless camera 2D measurements \\
\( \mathbf{X} \) & Lensless camera unknown scene intensity \\
\( \mathbf{y} \) & Flattened lensless camera measurements \\
\( \mathbf{x} \) & Flattened unknown scene intensity \\
\( h_I, w_I \) & Height and width of the lensless image \\
\( h_S, w_S \) & Height and width of the scene \\
\( \mathcal{C}(\cdot) \) & Cropping function \\
\( \mathcal{Z}(\cdot) \) & Zero padding function \\
\( \mathbf{h} \) & Pre-captured PSF \\
\( x, y \) & Display pixel coordinates \\
\( u, v \) & Light field angular coordinates \\
\( s, t \) & Image coordinates \\
\( \mathbf{L} \) & Light field \\
\( \mathbf{I}_{\text{pred}} \) & Predicted lensless capture \\
\( \mathbf{I}_{\text{gt}} \) & Ground truth lensless capture \\
\( \mathbf{F}_\theta \) & Multi-layer perceptron \\
\( \gamma(\cdot) \) & Positional encoding \\
\( n \) & Aperture size \\
\( m \) & Distance between display pixel and aperture \\
\( \alpha \) & Incident angle of the light rays \\
\hline
\end{tabular}
\caption{Notation used throughout this section.}
\label{tbl:notation}
\end{table}

\subsection{Lensless 2D Image Formation}\label{sec:lensless_formation}
Capturing the incoming light from pixels across multiple directions is crucial in our application. 
We choose the lensless cameras for their ease of modification and flexibility.
\begin{figure}[H]
  \includegraphics{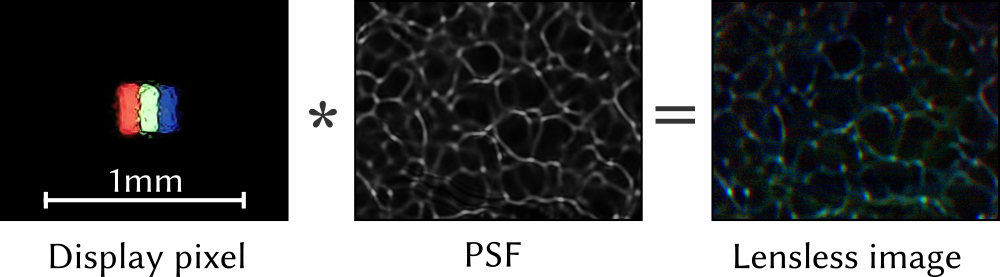}
  \caption{ The display pixel captured with a microscope (left). The \PSF captured with the proposed lensless camera (middle). The captured lensless image from the display pixel (right).}
  \label{fig:lensless_convolution}
\end{figure}
Our lensless camera contains a diffuser that alters the light propagation paths, resulting in distorted images on the sensor as shown in \refFig{lensless_convolution}.  
Let \( h_I \) and \( w_I \) denote the height and width of the sensor, and \( h_S \) and \( w_S \) denote the height and width of the scene.  
The forward imaging process can be expressed as  
\begin{equation}
  \mathbf{y}_I = \mathbf{H} \mathbf{x}_S,
\end{equation}
where \( \mathbf{y}_I \in \mathbb{R}^{(h_I \cdot w_I) } \) denotes the flattened measurement vector from the sensor,  
\( \mathbf{H} \in \mathbb{R}^{(h_I \cdot w_I) \times (h_S \cdot w_S)} \) is a coefficient matrix modeling the input light's interaction with the diffuser,  
and \( \mathbf{x}_S \in \mathbb{R}^{(h_S \cdot w_S)} \) represents the flattened intensity vector of the unknown scene.  
However, reconstructing the full \( \mathbf{H} \) matrix is computationally intractable.  
To address this, we position the diffuser close to the sensor to ensure spatial invariance in the forward model,  
allowing it to be approximated as a linear convolution:
\begin{equation}
  \begin{aligned}
    \mathbf{Y} 
    &= \mathcal{C}\big( \mathcal{Z}(\mathbf{X}) * \mathcal{Z}(\mathbf{h}) \big) \\
    &= \mathcal{C}\left( 
        \mathcal{F}^{-1} \left( 
          \mathcal{F} \big( \mathcal{Z}(\mathbf{X}) \big)
          \cdot
          \mathcal{F} \big( \mathcal{Z}(\mathbf{h}) \big)
        \right)
      \right),
  \end{aligned}
  \label{eq:formation_1}
\end{equation}
where $\mathbf{Y} \in \mathbb{R}^{h_I \times w_I}$ denotes the 2D measurements,  $\mathbf{X} \in  \mathbb{R}^{h_S \times w_S}$ denotes the unknown scene intensity, $*$ denotes the 2D linear convolution operation, $\mathcal{C}(\cdot)$ is the cropping operation, $\cdot$ is the Hadamard product, $\mathcal{Z}(\cdot)$ is the zero padding operation to ensures dimension consistency and avoids circular convolution artefact, $\mathcal{F}$ and $\mathcal{F}^{-1}$ denote the \DFT and \IDFT, respectively, and $\mathbf{h} \in \mathbb{R}^{h_S \times w_S}$ is the pre-captured \PSF.
To reconstruct for the unknown scene $\mathbf{X}$, we need to solve an inverse problem of the ~\refEq{formation_1} by minimizing the loss:
\begin{equation}
  \hat{\mathbf{X}} \gets \underset{\mathbf{X}}{\mathrm{argmin}} \; \mathcal{L}\big(\mathbf{Y}_{\text{gt}}, \mathcal{C}(\mathcal{Z}(\mathbf{X}) * \mathcal{Z}(\mathbf{h}))\big),
  \label{eq:minimization_0}
\end{equation}
where $\mathbf{Y}_{\text{gt}}$ is the ground truth lensless image.

\subsection{Lensless Light Field Formation}
We leverage the lensless camera system to capture the unknown light fields for pixels at different locations on the display $\mathbf{L} = (u,v,s,t)$, where the $(u,v)$ and $(s,t)$ are the angular and spatial coordinates. 
An image formed by the directional light rays from a single or multiple light sources satisfies:
\begin{equation}
  \mathbf{I} \left( s, t \right)=\int_{\Omega_x} \int_{\Omega_y} \mathbf{L} \left( u, v, s, t \right) d u \, d v,
\end{equation}
where $\{\Omega_x, \Omega_y\}$ is a subset of all the angles which the light rays reach for the imaging sensor.
To simplify the problem, we model the phase mask as an array of pinholes that scatter incident light rays arriving at angles $(u^\prime, v^\prime)$ onto the imaging sensor, forming distinct patterns that we refer as sub-aperture images.
The formation of each sub-aperture image can be modeled as the 2D convolution of the scene information with its corresponding \PSF patch, $\mathfrak{h}(u^\prime, v^\prime)$.
To compute this convolution efficiently across all angular dimensions, we employ a 4D \DFT on the spatial coordinates:
\begin{equation}
  \mathbf{I}_{\text{\text{gt}}} = 
  \mathcal{C}  
    \left( 
      \mathcal{F}^{-1} \left( 
        \mathcal{F} \big( 
          \mathcal{Z}(\mathbf{L}) 
        \big) 
        \cdot \mathcal{F} \big( 
          \mathcal{Z}(\mathfrak{h}) 
        \big) 
      \right) 
    \right).
\end{equation}
Finally, we pose the recovery of the light field as an inverse problem:
\begin{equation}
  \hat{\mathbf{L}} \gets \underset{\mathbf{L}}{\mathrm{argmin}} \; \mathcal{L}\Big( \mathbf{I}_{\text{\text{gt}}}, \mathcal{C} \big( \mathcal{Z}(\mathbf{L}) * \mathcal{Z}(\mathfrak{h}) \big) \Big),
  \label{eq:minimization_1}
\end{equation}
where $\mathbf{I}_{\text{gt}}$ denotes the ground truth lensless capture.

\subsection{Implicit Neural Representation}\label{sec:inr}
\begin{figure}
  \includegraphics{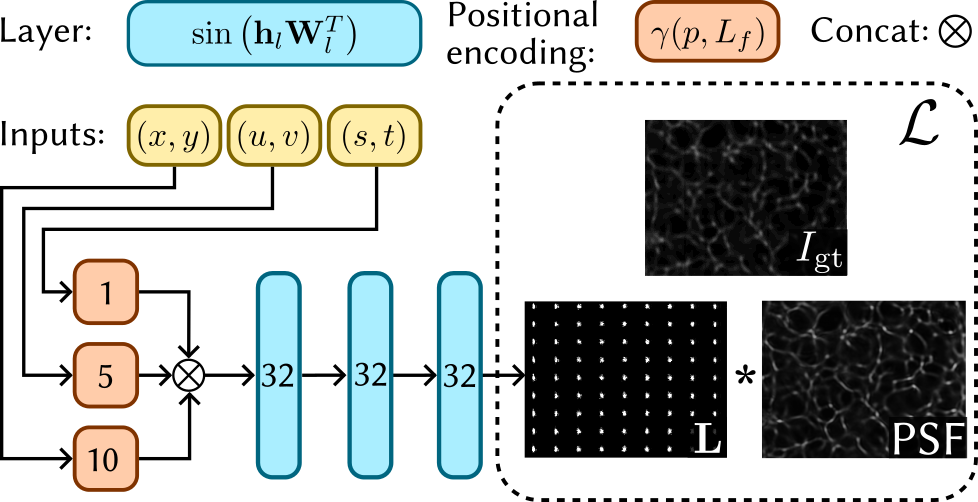}
  \caption{
  Each \MLP layer consists of 32 neurons and a sinusoidal activation function.
  We apply positional encoding with varying frequency levels ($L_f$) to each input coordinate group.
  After concatenating these encoded features, the model processes them to reconstruct the light field.
  We then perform linear convolution between this reconstruction and the pre-captured \PSF to generate the predicted lensless image ($\mathbf{I}\text{pred}$).
  Finally, we compute the loss with $\mathbf{I}\text{pred}$ and $\mathbf{I}_\text{gt}$.
  }
  \label{fig:model_pipeline}
\end{figure}
We utilize an \MLP to efficiently represent the continuous pixel light field samples from the lensless captures to avoid extensive sampling on the target screen, which satisfies:
\begin{equation}
  \hat{\mathbf{L}}(\mathbf{x},\mathbf{y},u,v,s,t) = \mathbf{F}_{\theta}(\mathbf{x},\mathbf{y},u,v,s,t),
  \label{eq:inr}
\end{equation}
where $\mathbf{F}_{\theta}$ is an \MLP with parameters $\theta$.
This network takes the spatial and angular coordinates as input and outputs the corresponding pixel color value.
The variables $\mathbf{x}$ and $\mathbf{y}$ denote the illuminated pixel coordinates on the display. 
Before the input coordinates $p$ are fed into the \MLP, we apply positional encoding~\cite{mildenhall2021nerf} to preserve high-frequency details:
\begin{equation}
\gamma( p )=\left( \, \sin\bigl( 2^{0} \pi p \bigr), \, \cos\bigl( 2^{0} \pi p \bigr), \, \cdots, \, \sin\bigl( 2^{L-1} \pi p \bigr), \, \cos\bigl( 2^{L-1} \pi p \bigr) \, \right),
\end{equation}\label{eq:encoding}
We apply the positional encoding with different frequency levels in the input coordinate groups: $\mathbf{e} = \bigl(\gamma_0(\mathbf{x}, \mathbf{y}), \gamma_1(u,v), \gamma_1(s,t) \bigr)$.
Finally, the minimization problem in \refEq{minimization_1} is reformulated as:
\begin{equation}
  \hat{\theta} \gets \underset{\theta}{\mathrm{argmin}} \; \mathcal{L}\Big( \mathbf{I}_{\text{gt}}, \mathcal{C} \big( \mathcal{Z}(\mathbf{F}_{\theta}(\mathbf{e})  ) * \mathcal{Z}(\mathfrak{h}) \big) \Big),
  \label{eq:minimization_2}
\end{equation}
where $\hat{\theta}$ denotes the optimized parameters of the \MLP.
\paragraph*{Loss function.}
We employ the $L_1$ norm to quantify the discrepancy between 
$\mathbf{I}_{\text{pred}}$ and $\mathbf{I}_{\text{gt}}$, 
and use $\mathcal{R}$ to penalize predicted pixel values exceeding the range $[0,1]$.
We define:
\begin{equation}
\mathcal{R}(\mathbf{I}_{\text{pred}}) =
\sum \max(\mathbf{I}_{\text{pred}} - 1, 0) +
\sum \max(-\mathbf{I}_{\text{pred}}, 0).
\label{eq:range}
\end{equation}
The total training loss is then:
\begin{equation}
\mathcal{L} =
\lambda_0 \|\mathbf{I}_{\text{gt}} - \mathbf{I}_{\text{pred}}\|_1 +
\lambda_1 \mathcal{R}(\mathbf{I}_{\text{pred}}),
\label{eq:loss}
\end{equation}
where $\lambda_0$, and $\lambda_1$ represent weights
($\lambda_0 = 1$,  $\lambda_1 = 10^{-7}$).

\section{Evaluation and Discussion}
\begin{figure}
  \includegraphics{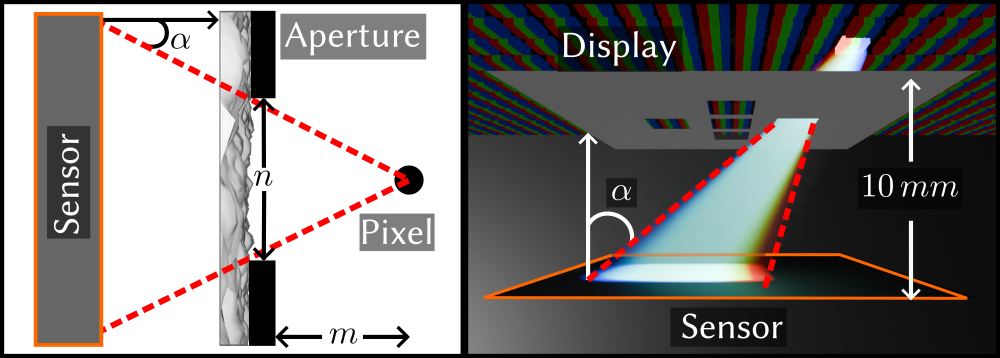}
  \caption{
  The incident angle ($\alpha$) of the light rays is determined by the distance between the aperture and the display pixel ($m$), as well as the size of the aperture ($n$) (left).
  Our proposed aperture array expands the incident angle range by turning on each pixel one by one (right).
  }
  \label{fig:incident_angle}
\end{figure}

\begin{figure}
  \includegraphics{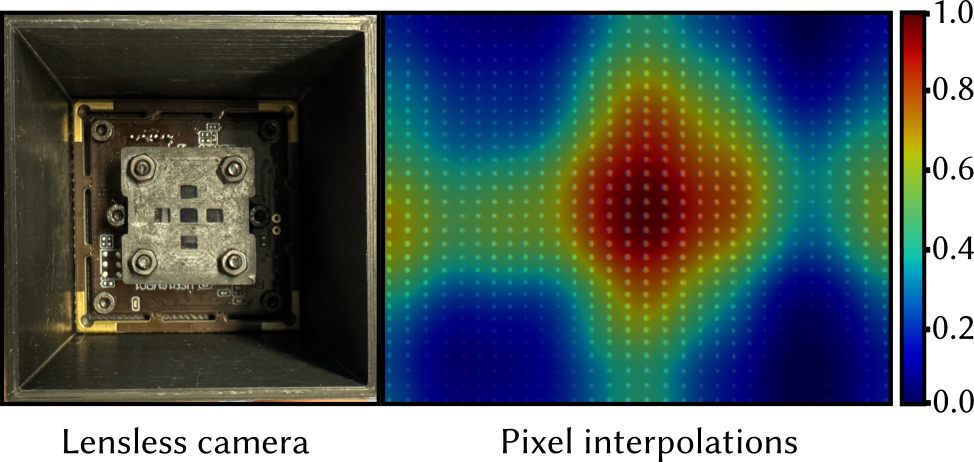}
  \caption{The top view of our lensless camera (left).
  The simulated display pixels with unseen incident angles and the corresponding overlay intensity heatmap (right).}
  \label{fig:rendered_display}
\end{figure}

\paragraph*{Hardware.}
Our hardware uses a $0.5^\circ$ engineered diffuser (Edmund 35-860) with five $2.5\,mm \times3\,mm$ apertures positioned $10\,mm$ from the imaging sensor as depicted in \refFig{incident_angle}.
We experimentally determine this spacing to ensure spatial invariance and sharp caustic patterns, satisfying \refEq{formation_1}.
Moreover, we place the apertures such that we maximize the angle of rays arriving from a pixel landing on an imaging sensor.
This aperture placement maximizes ray angles from pixels to the sensor, though sensor dimensions limit the maximum supported angle.
In our implementation, the limited imaging sensor dimensions dictate the locations of these apertures, posing a constrain on the maximum angle we can support with our aperture arrangement.
We 3D print a housing maintains $30\,mm$ between diffuser and screen, as demonstrated in left image \refFig{rendered_display}.
We capture a stack of \PSF for each aperture using a white LED with a $100\,\mu m$ pinhole (Thorlabs P100HKb), then mount the prototype in front of the display and activate pixels sequentially.
To account for \LCD backlight bleeding~\footnote{https://pixiogaming.com/blogs/latest/understanding-backlight-bleed-in-ips-panels-causes-and-solutions}, we sample nine screen positions to capture the full spatial distribution (Supplementary Sec. 1).
\paragraph*{Implicit Neural Representation.}
We divide lensless captures into $9 \times 9$ sub-aperture images using a $54 \times 70$ pixel sliding window.
Later, we feed them into a \MLP that consists of 3 fully connected layers with sinusoidal activation functions and 32 channels each.
To improve generalization, we inject random noise into the coordinates from \refSec{inr}: display (std: $5 \times 10^{-3}$), angular (std: $1 \times 10^{-2}$), and subview (std: $1 \times 10^{-3}$).
We then apply positional encoding \refEq{encoding} with 1, 5, and 10 frequency levels for display, angular, and spatial coordinates, respectively (see Fig.~\ref{fig:model_pipeline} and Supplementary Sec. 2).
We optimize using Adam optimizer with an initial learning rate of $0.001$ that decays linearly over $800$ epochs and clip gradient norms to $1.0$.
We train the model on independent color channels to mitigate crosstalk, which takes approximately one hour on an NVIDIA RTX 2070 GPU, while inference takes $0.01$ seconds per 1080p frame.
\paragraph*{Quantitative Evaluations.}
To assess pixel consistency, we measure intensity variations in reconstructed light fields across continuous horizontal and vertical incident angles.
We sample nine display positions with five measurements at different apertures per position, train the \MLP on this data, and estimate pixel values at unseen angles.
\refFig{rendered_display} shows smooth predicted intensity changes, demonstrating the model's angle-dependent reconstruction capability.
Our architecture outperforms a vanilla \MLP baseline with fully connected ReLU layers across all metrics (\refTbl{comparison}).
\begin{figure}
  \includegraphics{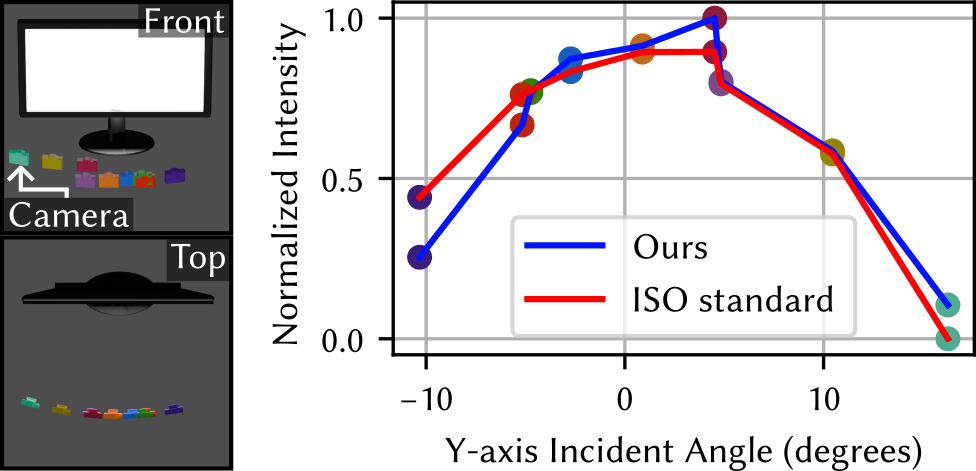}
  \caption{
  The illustration of the camera poses and the display (left).
  The average intensity that normalized to $[0,1]$ from the ISO standard and ours, with data samples color-coded by the corresponding camera poses  (right).
  }
  \label{fig:ours_vs_conventional_plot}
\end{figure}
To benchmark our method, we follow the ISO standard~\cite{ISO9241-305}: align a camera with the display in a dark room, display full-screen white stimuli, and capture images from $-10^\circ$ to $16^\circ$ vertical incident angles.
\refFig{ours_vs_conventional_plot} shows our method reproduces the ISO intensity trend, validating its physical plausibility.
We capture light field data from nine display positions without camera rotation or controlled lighting, reducing measurement time and simplifying view-dependent calibration while maintaining comparable accuracy.
\begin{table}[h]
\centering
\caption{Model Comparison}
\begin{tabular}{c|c|c|c|c|c}
\hline
\textbf{Methods} & \makecell{\textbf{PSNR}\\(dB)} $\uparrow$ & \textbf{SSIM} $\uparrow$ & \textbf{MSSIM} $\uparrow$ & \makecell{\textbf{Train}\\(h)} & \makecell{\textbf{Inference}\\(s)}  \\
\hline
Ours & \textcolor{darkgreen}{\textbf{19.54}} & \textcolor{darkgreen}{\textbf{0.9165}} & \textcolor{darkgreen}{\textbf{0.9549}} & \textcolor{darkred}{1} & \textcolor{darkred}{0.01} \\
Vanilla & \textcolor{darkred}{9.93} & \textcolor{darkred}{0.5327} & \textcolor{darkred}{0.8049} & \textcolor{darkgreen}{0.8} & \textcolor{darkgreen}{0.003} \\
\hline
\end{tabular}
\label{tbl:comparison}
\end{table}

\paragraph*{Limitations and Future Works.}
Our $46.6^\circ$ angular coverage remains well below the $240^\circ$ required in professional display calibration.
We could extend this range by co-optimizing the apertures and the diffuser design.
Beyond angular coverage, pixel-wise training hinders scalability for full-panel characterization.
To overcome these constraints, we outline several promising research directions.
First, developing an end-to-end pipeline that eliminates \PSF convolutions, cropping, and padding.
Second, hash encoding~\cite{muller2022instant} or Gaussian splatting could model light fields more efficiently.

\appendix

\bibliographystyle{ACM-Reference-Format}
\bibliography{references.bib}

\end{document}